\documentclass{article}

\PassOptionsToPackage{numbers}{natbib}
\usepackage[preprint]{neurips_2021}

\usepackage[utf8]{inputenc} % allow utf-8 input
\usepackage[T1]{fontenc}    % use 8-bit T1 fonts
\usepackage[colorlinks,citecolor=gray,linkcolor=purple]{hyperref}       % hyperlinks
\usepackage{url}            % simple URL typesetting
\usepackage{booktabs}       % professional-quality tables
\usepackage{amsmath}
\usepackage{amsfonts}       % blackboard math symbols
\usepackage{nicefrac}       % compact symbols for 1/2, etc.
\usepackage{microtype}      % microtypography
\usepackage{xcolor}         % colors
\usepackage{xspace}
\usepackage{enumitem}
\usepackage{graphicx}
\usepackage{overpic}

\hypersetup{bookmarksdepth=2}

\renewcommand{\paragraph}[1]{\noindent\textbf{#1}~~}

\newcommand*{\ours}{ViPL4S}
\newcommand*{\waymo}{\textsc{waymo}\xspace}
\newcommand*{\rexk}{\textsc{re{\small10}k}\xspace}

\newcommand*{\eg}{e.g.\@\xspace}
\newcommand*{\ie}{i.e.\@\xspace}

\title{Unsupervised Video Prediction from a Single Frame by Estimating 3D Dynamic Scene Structure}

\author{%
  Paul Henderson\\
  \texttt{paul@pmh47.net}\\
   \And
  Christoph H. Lampert~\thanks{Equal advising}\\
  \texttt{chl@ist.ac.at}\\
   \And
  Bernd Bickel~\footnotemark[1]\\
  \texttt{bernd.bickel@ist.ac.at}\\
  \And \\[-18pt]
  Institute of Science and Technology (IST) Austria\\ \\
    \url{http://pmh47.net/vipl4s/}
}

\begin{document}

\maketitle

\begin{abstract}
  Our goal in this work is to generate realistic videos given just one initial frame as input.
  Existing unsupervised approaches to this task 
%  rely on \pmh{large / unstructured / uninterpretable / black-box / monolothic} deep generative models \pmh{whose latent variables are XXX}
%  ; they therefore 
  do not consider the fact that a video typically shows a 3D environment, and that this should remain coherent from frame to frame even as the camera and objects move.
  We address this by developing a 
%  \pmh{In contrast, we present a }
  model that first estimates the latent 3D structure of the scene, including the segmentation of any moving objects.
  It then predicts future frames by simulating the object and camera dynamics, and rendering the resulting views.
  Importantly, it is trained end-to-end using only the unsupervised objective of predicting future frames, without any 3D information nor segmentation annotations.
  Experiments on two challenging datasets of natural videos show that our model can estimate 3D structure and motion segmentation from a single frame, and hence generate plausible and varied predictions.
\end{abstract}

\section{Introduction}
\label{sec:introduction}

Predicting the future from a single image is a compelling problem.
As humans, we can readily imagine what might happen next in the scenario depicted in an image---which objects are moving, which are static, where the camera might move to.
Indeed, we can visualize different possible futures, \eg with a particular car being either parked or driving, when this is uncertain given just a single image.
However, this task remains challenging for machine learning methods, as the predicted video is very high-dimensional, yet must retain temporal coherence over a number of frames.

%\pmh{The ambiguity of the problem---or more precisely, the stochasticity of the prediction task---makes it an extremely challenging one, particularly in the high-dimensional space of realistic videos.}

Most real-world videos are filmed by a camera moving through a dynamic 3D environment or \textit{4D scene}.
This scene is typically stable, in the sense that many aspects remain constant---for example, the shape of a car or the color of a dog do not change over time, even though the car, dog and camera may be moving.
State-of-the-art unsupervised methods for future prediction cannot exploit this powerful prior knowledge.
They rely on black-box generative models~\cite{denton18icml,franceschi20icml,kumar20iclr}, that must instead try to learn it from data.
In practice, they often fail to do so, with objects slowly changing in shape or texture over time~\cite{villegas19neurips}.

%\pmh{existing models often have a latent dynamics model, but doesn't reflect world structure}

\begin{figure}
  \centering
  \includegraphics[width=0.9\linewidth]{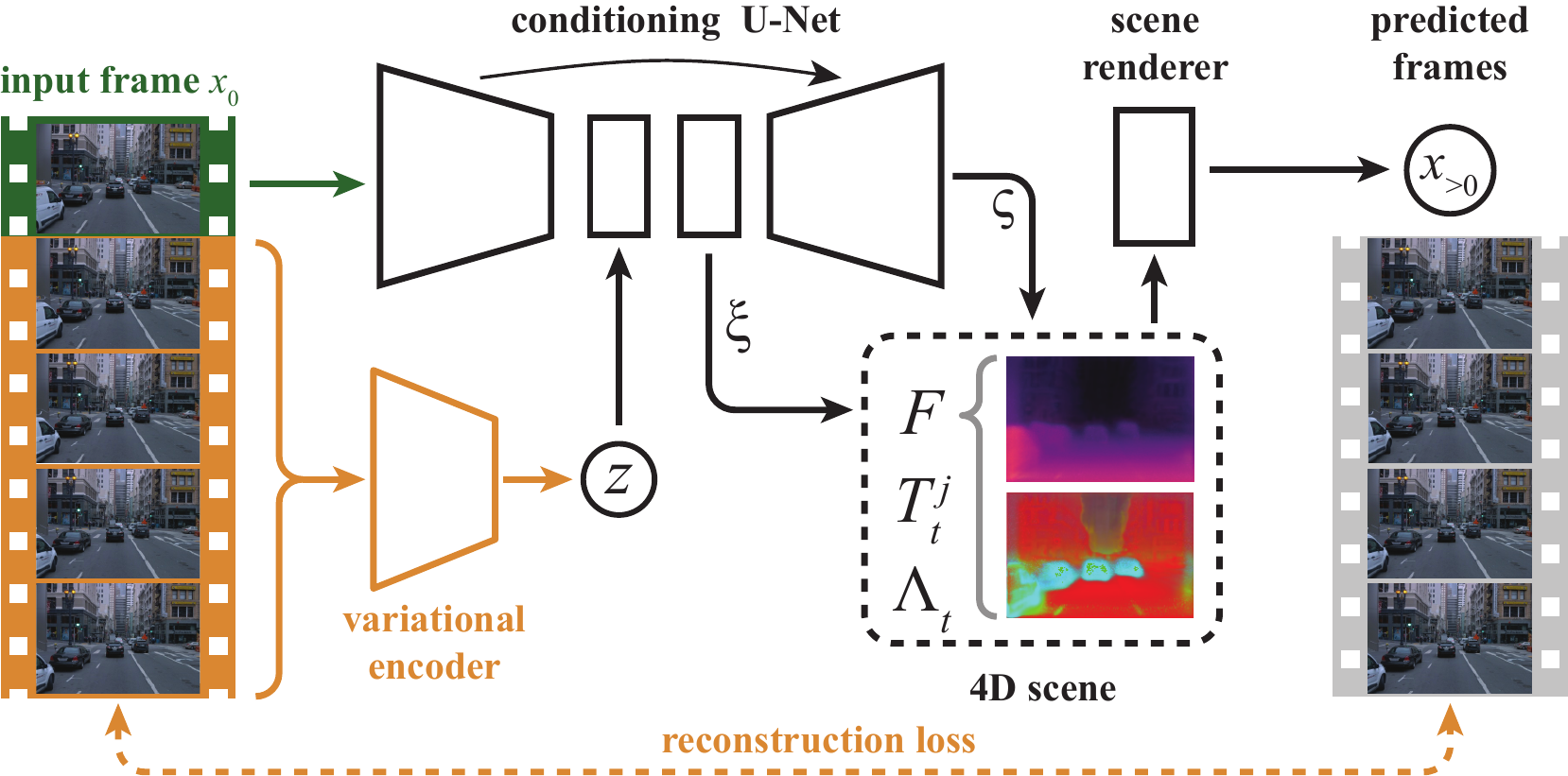}
  %\vspace{-6pt}
  \caption{Overview of our conditional generative model of video. Given a single frame $x_0$, our model learns to generate plausible video clips $x_{>0}$. We decode $x_0$ and a latent Gaussian embedding $z$ to give a 4D scene, consisting of the 3D structure and motion segmentation (represented by a function $F$), camera motion $\Lambda_t$, and transformations of moving entities $T_t^j$. Our method is trained like a VAE (orange) to reconstruct input frames via $z$ and the 3D scene representation. 
%   \pmh{illustrate F taking xyz and yielding rho etc.?} 
%   \pmh{add flow/keypoint loss!} 
%   \pmh{add daml between frames and networks?} 
%   \pmh{explicitly label `latent 3d geometry' and `motion segmentation'}
  }
  \label{fig:model}
\end{figure}

We propose a novel approach to video prediction from a single frame, that first estimates the 3D structure of the scene depicted in that frame---its 3D geometry, texture, and segmentation into moving entities (Section~\ref{sec:method}).
We assume that the camera and other entities in the scene move, but retain their original appearance and geometry; this allows us to generate future frames by predicting their motion, and re-rendering the updated scene.
%We then predict future frames by assuming that the camera and other entities in the scene move, but otherwise retain their original appearance and geometry.
%\pmh{Given this, we then define future frames as renderings of the scene}
%
Of course, learning to predict 3D structure and motion segmentation from a single image is a challenging problem in itself, with state-of-the-art methods relying on extensive supervision~\cite{ranftl20pami,ma19cvpr,he17iccv}.
Here we adopt an entirely \textit{un}supervised approach, that requires only monocular RGB videos during training.
\textbf{Our model is trained end-to-end for video prediction, treating the 4D scene structure as a latent factor to be estimated.}
We adopt a variational approach that handles the inherent stochasticity of the problem, and is able to sample different possible structures and future motion for a given frame.
%\pmh{Concretely, our model embeds all uncertainty (that is not fixed given the initial frame) about the structure and future motion in a latent variable}
%\pmh{unlike existing models that map frame and z directly to pixels, we decode via an explicit latent representation of the dynamic scene; this repr is novel!}
%\pmh{...and explain sequences in terms of it}

As a key ingredient in our model, we develop a novel representation of 4D scenes, that factors geometry and motion information.
We build on the static scene representation of \cite{mildenhall20eccv}, modifying it to represent the segmentation of a scene into comoving regions.
This allows us to use a low-dimensional, factorized motion representation, specifying time-varying transforms for each of a small number of motion components, rather than a dense flow field for the entire scene.
It also directly captures the prior knowledge that many points in a scene typically move together (\eg all the points within an object), without the need for extra regularization.
%
%\pmh{compare with o3v and scalor---`globalness' of components}

We conduct a comprehensive evaluation of our method on two datasets of challenging real-world videos (Section~\ref{sec:experiments}).
Unlike most datasets used in prior works~\cite{denton18icml,lee18arxiv,franceschi20icml}, these include (unknown) 3D camera motion and complex appearances.
We show that our method successfully recovers scene structure, and can use this to generate diverse and plausible predictions of future frames.
Moreover, its quantitative performance significantly exceeds an existing state-of-the-art work~\cite{franceschi20icml}.

%\pmh{somewhere emphasise that most prior works only consider the multiple-input-frames setting}

\paragraph{Contributions.}
In summary, our contributions are as follows:
\begin{itemize}[itemsep=1pt,leftmargin=16pt,topsep=0pt]
    \item we propose a novel representation for multi-object 4D scenes
    \item we show how to incorporate this in a conditional generative model of videos, that can be trained without supervision
    \item ours is the first unsupervised video prediction method that ensures predicted frames represent a coherent 4D scene
\end{itemize}

\section{Method}
\label{sec:method}

%Following \cite{denton18icml,franceschi20icml} \pmh{cite others?}, 
We cast the problem of stochastic video prediction as drawing samples from a conditional generative model (Figure~\ref{fig:model}).
We use $x_0$ to denote the initial frame that is provided as conditioning.
To define the conditional distribution $P(x_1,\, \ldots,\, x_L \,|\, x_0)$ on $L$ subsequent frames $(x_1,\, \ldots,\, x_L) \equiv x_{>0}$, we introduce a latent variable $z$ that will capture all uncertainty about the future, \ie embed all information that is not present in $x_0$.
This will include the camera and object motion, and the appearance of regions disoccluded by moving objects.
This leads to a probabilistic model of the form $P(x_{>0} \,|\, x_0)=P(z)\,P(x_{>0} \,|\, z ,\, x_0)$.
As is common~\cite{kingma14iclr,rezende14icml}, we choose $z$ to be a multivariate standard Gaussian distribution, \ie $z \sim \mathcal{N}(\mathbf{0},\, I)$.
To ensure the generated clip depicts a consistent scene,
the conditional distribution on future frames $P(x_{>0} \,|\, z ,\, x_0)$ is given by first mapping $x_0$ and $z$ to a description $\mathcal{S}$ of a 4D scene, incorporating geometry, texture, segmentation, and motion (dashed box in Figure~\ref{fig:model}).
Then, the frames $x_t$ for $t>0$ are generated by \textit{rendering} $\mathcal{S}$ at the relevant times $t$, and finally adding Gaussian pixel noise to ensure the data likelihood is always defined.
Note the contrast with earlier works~\cite{denton18icml,lee18arxiv,franceschi20icml} that predict $x_{>0}$ directly from $x_0$ and $z$, \eg using a CNN, and without the intermediate representation $\mathcal{S}$ that ensures a consistent scene is shown.

We first describe a recently-proposed static scene representation~\cite{mildenhall20eccv} and how it is rendered to pixels; we then extend it to give our 4D scene representation $\mathcal{S}$.
Next, we discuss how our representation is conditioned on $x_0$ and $z$, to allow inferring it from a single frame and incorporating it into our probabilistic model.
Finally, we describe how the overall model is trained end-to-end.

\paragraph{Background: NeRF representation of static scenes.}
\textit{Neural radiance fields} (NeRFs)~\cite{mildenhall20eccv} represent a single static scene using a function $F$ that maps points $p \in \mathbb{R}^3$ in the 3D space of the scene to a density $\rho$ and an RGB color $c \in [0,\,1]^3$.
The density $\rho \in \mathbb{R}^+$ represents the differential probability that a light ray cast from the camera will terminate (\ie hit some matter) at point $p$; zero corresponds to free space, and large values to solid objects.
In practice $F$ is a densely-connected neural network, with $p$ preprocessed using the Fourier embedding method of \cite{tancik20neurips}.
To render the scene defined by $F$, standard quadrature-based volume rendering techniques~\cite{max95tvg,mildenhall20eccv,martinbrualla21cvpr}, are used.
%
% INTEGRAL-ISH VERSION, SIMILAR TO PIXELNERF; unclear how to incorporate cpt-weights
% \begin{equation}
%     \int_{0}^{\infty} H(\lambda) \, \rho(r(\lambda; q)) \, c(r(\lambda; q)) \, d\lambda
% \end{equation}
% where
% \begin{equation}
%     H(\lambda) = \exp\left( - \int_0^\lambda \rho(r(\mu; q)) \, d\mu \right)
% \end{equation}
% and
% $r(\lambda; q)$ gives the 3D point at distance $q$ along the ray cast from the camera at pixel $q$. \pmh{clarify}
%
Specifically, to calculate the color $C(q)$ of the pixel at coordinate $q$, we evaluate $F$ at $K$ points along the ray cast from $q$ into the scene through the camera lens, setting
%
% SUM-ISH VERSION, SIMILAR TO NERF-W
\begin{equation}
\label{eq:nerf}
    C(q) = \sum_{k=1}^{K} V_k \left( 1 - \exp(-\rho(r_k)\delta_k) \right) c(r_k) ,
    \hspace{8pt}\text{where}\hspace{8pt}
    V_k = \exp\left( - \sum_{k'=1}^{k-1} \rho(r_{k'}) \delta_{k'} \right) ~~~
\end{equation}
Here
$r_k$ gives the $k$\textsuperscript{th} 3D sample point along the ray from $q$, and $\delta_k=||r_{k+1}-r_k||$ is the spacing between samples.
Intuitively, $V_k$ represents how visible the $k$\textsuperscript{th} sample is in spite of occlusion by nearer samples.

\paragraph{Our 4D scene representation.}
We represent a 4D scene $\mathcal{S}$ as three parts: (i) the static scene geometry and appearance at $t=0$; (ii) its segmentation into $J$ different \textit{motion components}; and (iii) a set of time-varying transformations $T_t^{1 \ldots J}$ capturing these components' motion for $t>0$.
Intuitively, each motion component is a region of 3D space in the scene, in which all the points are moving coherently.
%Note that we allow arbitrary points to be assigned to each motion component---
The model is free to use motion components to represent individual moving objects (\eg a car), comoving groups of objects (\eg a convoy of cars), or the static background.
The transformation $T_t^j$ defines how points in the initial scene belonging to component $j$ are mapped to their location in frame $t$.
%
%\pmh{thus imply a different distribution/placement of density + color in the later frames} \pmh{as the choice of components is a soft assignment, the density is actually spread to different locations!}
%
%To represent the initial scene geometry and color, we take the approach described above---a function $F(p)$ maps points in the 3D scene at $t=0$ to a density $\rho$ and color $c$.
To implement this, we introduce a \textit{scene function}, $F(p)=[\rho(p),c(p),\omega(p)]$; here $\rho$ and $c$ are the density and color at point $p$ in the 3D scene corresponding to $t=0$, similar to above.
In order to capture the fact that different regions of the scene belong to different motion components, $F$ also outputs a vector $\omega \in [0,\,1]^J$ of values indicating which of $J$ components each point belongs to; these are normalized by a softmax.
Intuitively, if $\omega$ is one-hot with $\omega_j(p)=1$, then the point $p$ will transform according to $T^j_t$; if $\omega$ is not one-hot, then corresponding fractions of the density at $p$ will transform according to each $T^j_t$.
%
%\pmh{explanatory figure?}

Together, the scene function $F$ and component transformations $T_t^{1 \ldots J}$ define the distribution of density and color in all frames $t=1 \ldots L$.
To render frame $t$, we adapt (\ref{eq:nerf}) to our dynamic multiple-component setting.
Suppose that the camera transformation (\ie extrinsic matrix, representing its location and rotation) in frame $t$ is denoted by $\Lambda_t$.
%To render the scene defined by $F$ and $T_t^{1 \ldots J}$ for each frame $t=1 \ldots T$, we adapt (\ref{eq:nerf}) to our dynamic multiple-component setting.
We calculate the density and color for point $p$ at time $t$ by applying the inverse of the component transformations, to find which location $p_0^j$ in the initial scene would be transported to $p$ by the $j$\textsuperscript{th} component's motion, \ie $p_0^j = (T_t^j)^{-1}[p]$.
Compared with (\ref{eq:nerf}), we must now evaluate $F$ once per motion component for each sample; this effectively defines one opacity per component, and we weight these according to $\omega_j$, before summing to give the combined density and color.
Therefore, the expected color $x_t(q)$ of the pixel at coordinate $q$ in predicted frame $x_t$ is given by
%
% SUM-ISH AND WITH CPTS/MOTION
\begin{equation}
    x_t(q) = \sum_{k=1}^{K} \left\{ 
        V_k \, 
        \sum_{j=1}^J \left[ 
%            \omega_j((T_t^j)^{-1}[r^t_k]) \left( 1 - \exp(-\rho((T_t^j)^{-1}[r^t_k])\delta_k) \right) \, c((T_t^j)^{-1}[r^t_k])
            \alpha_k^j \, \omega_j\big((T_t^j)^{-1}[r^t_k]\big) \, c\big((T_t^j)^{-1}[r^t_k]\big)
        \right]
    \right\}
\end{equation}
where
\begin{equation}
%    V_k = \exp\left( - \sum_{k'=1}^{k-1} \sum_{j=1}^J \rho((T_t^j)^{-1}[r^t_{k'}]) \, \omega_j((T_t^j)^{-1}[r^t_{k'}]) \delta_{k'} \right)
    V_k = \exp\left(\sum_{k'=1}^{k-1} \log \prod_{j=1}^J (1 - \alpha_k^j) \, \omega_j\big((T_t^j)^{-1}[r^t_{k'}]\big) \delta_{k'} \right)
\end{equation}
and
\begin{equation}
    \alpha_k^j = 1 - \exp\big( -\rho((T_t^j)^{-1}[r^t_k]) \, \delta_k \big)
\end{equation}
Here $r^t_k$ gives the $k$\textsuperscript{th} 3D sample point along the ray cast from the camera at pixel $q$ of frame $t$; this has an implied dependence on the camera transformation $\Lambda_t$.
%We denote the camera transformation (\ie extrinsic matrix, representing location and rotation) in each frame by $\Lambda_t$.
Note that $\omega$ weights the sample opacities $\alpha_k^j$, (\ie is applied \textit{after} integrating the densities) as these lie in $[0,1]$.
Although more intuitive, directly weighting the density $\rho$ has no practical effect, as $\rho$ is unbounded hence arbitrarily-large density may be assigned to all components, rather than the desired outcome of $\omega$ `splitting' a finite density between components.

\paragraph{Conditioning the scene representation.}
%
%\pmh{maybe re-link $F$ back to $c$ etc above}
The scene function $F$ and transformations $T_t^{1 \ldots J}$ and $\Lambda_t$ corresponding to a given input frame $x_0$ must be inferred from that frame and the latent $z$.
We therefore introduce a U-Net~\cite{ronneberger15miccai} \textit{conditioning network}\footnote{All network architectures are given in Appendix~\ref{app:architectures}}, that has $x_0$ as input, and injects $z$ as conditioning at the bottleneck layer; from this we extract a 2D feature map $\zeta$ and an embedding vector $\xi$.

$\xi$ is decoded by a densely-connected network to give parameters of the transformations $T_t^{1 \ldots J}$ and $\Lambda_t$ for the components and camera respectively.
We experiment with two different parametrizations of $\Lambda_t$: (i) a general model that allows arbitrary camera translation and yaw/pitch; (ii) a vehicle-specific model that captures the prior knowledge that cameras mounted on cars are typically restricted to motion parallel with the ground plane, characterized by a forward speed and azimuthal velocity.
We hypothesize this additional prior knowledge will make the learning task easier, as the model does not have to waste representational capacity learning it.
The precise specifications of these parametrizations are given in Appendix~\ref{app:camera-transforms}.

The conditioning variables $\zeta$ and $\xi$ also influence the scene function $F$ in two ways.
First, $\xi$ is used as input to FiLM conditioning layers~\cite{perez18aaai} following each hidden-layer activation~\footnote{We tried instead conditioning $F$ by having a \textit{hyper-network} directly predict its weights; this gave comparable results but proved to be much less stable during training}.
Second, when evaluating $F$ at a 3D location $p$ at time $t$, we map $p$ back into the 3D space of the initial frame ($t=0$) by applying the inverse of the predicted camera transformation and each of the component transformations at time $t$.
Then, we bilinearly sample the feature map $\zeta$, and provide these features as an additional input to $F$, concatenated with $p$, similar to \cite{yu21cvpr}.
%
%\pmh{in the above para, consider an equation! but if so need to make it clear what happens with the $j$ dependency}
%\pmh{in the above para, we seem to lose the fact that F has to be sampled once per component!}

\paragraph{Training and regularization.}
We have now specified the full conditional generative model $P(x_{>0} \,|\, x_0)$.
To train this model, we would ideally maximize the likelihood of future frames given initial frames.
However, this is intractable as we must marginalize the latent variable $z$; we therefore use Stochastic Gradient Variational Bayes~\cite{kingma14iclr,rezende14icml}.
Specifically, we introduce a variational posterior distribution $Q(z \,|\, x_0,\, x_{>0})$, chosen to be a diagonal Gaussian.
Its mean and log-variance are predicted by an encoder network $\mathrm{enc}_\phi(x_0, x_{>0})$.
For this we use a 3D CNN taking the concatenated frames as input (orange components in Figure~\ref{fig:model}).
Intuitively, as the encoder has access to complete video clips at training time, it should learn to embed information about motion and disoccluded regions in $z$, so the decoder can learn to reconstruct them.
In practice, the model is trained end-to-end to maximize the evidence lower bound (ELBO)~\cite{kingma14iclr,rezende14icml} using Adam~\cite{kingma15iclr}; as in \cite{higgins17iclr}, we multiply the KL-divergence term by a weight $\beta$, which is linearly increased from zero during the start of training.
We also add a reconstruction loss for $x_0$, with the component and camera transformations set to the identity.

To reduce computational intensity, we only reconstruct a sparse subset of points $q$ in each frame.
Moreover, to reduce the number of samples $K$ required along each ray, we also pretrain the unsupervised single-frame depth-prediction method of \cite{li2020corl} on our data, and use this to guide the sampling process at training time.
Specifically, we draw one third of samples nearer the camera than the depth prediction, one third in the region close to it, and one third farther away.
We also provide these depth predictions as an additional input to $F$ and regularize $\rho$ and $\Lambda_t$ to be consistent with them, which improves convergence.

%\paragraph{Regularization.}
%
To avoid degenerate solutions, we apply several other regularizers on the predictions of our model.
To do so, we render 2D flow and component segmentation maps $\mathcal{F}_t$ and $\mathcal{M}_t^j$ for each frame $x_t$.
Inspired by classical structure-from-motion techniques, we extract keypoint tracks from the ground-truth frames using the self-supervised method \cite{detone18cvpr}.
We then require that the frame-to-frame displacements of the keypoints are consistent with $\mathcal{F}_t$.
We also L1-regularize the component velocities, which avoids difficult-to-recover local optima where the motion is very large, and TV-L1 regularize~\cite{BoykovICCV01,li2020corl} the component masks $\mathcal{M}_t^j$.
Further details of these regularizers are given in Appendix~\ref{app:regularizers}.

\section{Related Work}

\paragraph{Stochastic multi-frame video prediction.}
Early works on video prediction did not account for the fact the the future is uncertain~\cite{ranzato14arxiv,finn16nips,debrabandere16nips}, hence tended to produce blurry or averaged predictions on real-world data~\cite{babaeizadeh18iclr,franceschi20icml}.
Recently there has been increasing work on \textit{stochastic} prediction methods, which do account for this uncertainty, usually with variational inference~\cite{kingma14iclr}.
These methods typically have one latent embedding variable per frame, with frames also conditioned on prior frames in an autoregressive fashion.
\cite{denton18icml} and \cite{he18eccv} use ConvLSTMs for the decoder and encoder, with the prior distribution on the latents also dependent on preceding frames. % MNIST,KTH,BAIR
\cite{villegas19neurips} present an extension using deeper networks to achieve improved visual fidelity. % KITTI, H36m, Towel-Pick robot
\cite{franceschi20icml} extend this further, with an ODE-inspired residual architecture to improve the dynamics model.
\cite{babaeizadeh18iclr} again use a variational approach, but with a more sophisticated generative process that composes output frames from transformed regions of the input frames.
\cite{lee18arxiv} augment the usual variational reconstruction loss with an adversarial term, to produce sharper results. % BAIR+KTH
Instead of variational training, \cite{kumar20iclr} propose an approach based on normalizing flows~\cite{rezende15icml}; this allows maximizing the exact likelihood, at the expense of an extremely computationally-heavy model.
Most of these methods are evaluated exclusively on datasets with a static camera~\cite{babaeizadeh18iclr,franceschi20icml,denton18icml}; moreover, the majority focus on multi-frame inputs---a considerably easier setting as object velocities can be observed.

\paragraph{Structured video prediction.}
A separate line of work aims to decompose videos into a set of objects, model their dynamics, and hence predict future frames. These are broadly similar in spirit to our model, but all consider objects as 2D entities rather than 3D, which limits their applicability to real-world datasets.
\cite{ye19iccv} predicts videos from a single frame, in terms of a set of moving objects defined by bounding-boxes; however they assume access to ground-truth boxes (or a pretrained detector) during both training and testing.
\cite{minderer19neurips} learns to auto-encode videos through a latent space that explicitly represents the locations of a set of keypoints, while also learning a dynamics model on that space.
\cite{kossen20iclr} construct a latent representation similar to early work on object-centric image generation \cite{eslami16nips}, but with a more-sophisticated dynamics model and the ability to generate conditioned on input frames.
\cite{lin20icml} learn a 2D sprite-based representation jointly with a dynamics model over those sprites; this allows video extrapolation on simple datasets that can be adequately modelled by 2D sprites.
Lastly, \cite{wu20cvpr,qi19cvpr} both predict frames by reasoning over optical flow (in 2D and 3D respectively), but assume access to additional information---depth images for \cite{qi19cvpr}, and segmentation masks for \cite{wu20cvpr}.
%
%\pmh{our neurips'20 cites a couple of others}

\paragraph{Structured video generation.}
Several recent methods build unconditional generative models of video, by modelling the video in terms of a set of objects.
\cite{jiang20iclr,crawford20aaai,veerapaneni20corl} model objects as 2D sprites, and focus on evaluating tracking and segmentation rather than generation.
\cite{henderson20neurips} focuses on generation, treating objects as 3D, but only considering short, synthetic videos.
These methods can sample videos \textit{a priori}, but are unable to condition on provided frames.

\paragraph{Related NeRFs.}
Our method uses a novel extension of NeRF~\cite{mildenhall20eccv} to represent 4D scenes.
\cite{pumarola20cvpr} present an alternative approach to this; however, they focus on the single-scene setting, \ie training one model to represent a single scene given multiple views---in contrast to our setting of inferring the representation from a single image.
In contrast, \cite{yu21cvpr} consider predicting a NeRF from one or few images, but only to represent a static scene.
%\pmh{someone german has a non-meta nerf on kitti/etc. scenes}
%\pmh{someone has a FiLM-conditioned nerf}

%\pmh{consider mentioning cvpr-ish supervised ones if we (as well as being less-supervised) can make stronger guarantees that the appearance/structure/etc. is `preserved'}

\begin{figure}[t]
    \centering
    \includegraphics[width=0.49\linewidth]{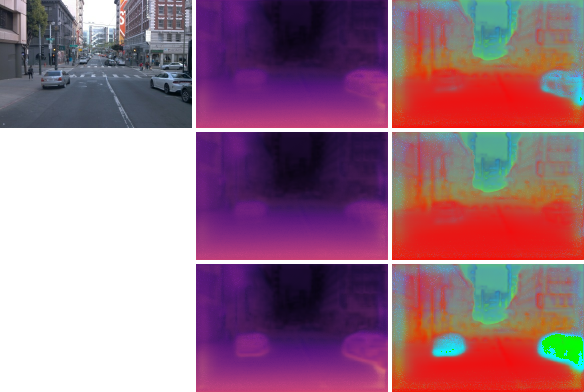}~
    \includegraphics[width=0.49\linewidth]{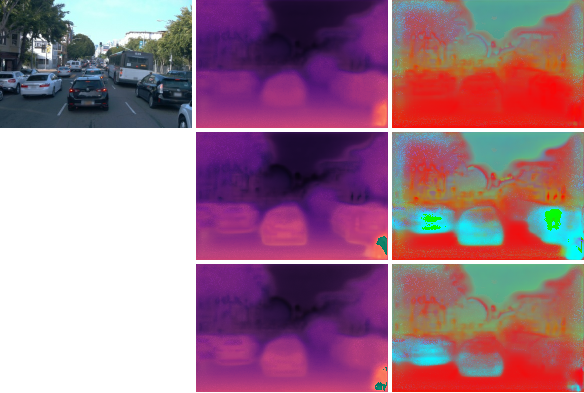}\\[4pt]
    \includegraphics[width=0.49\linewidth]{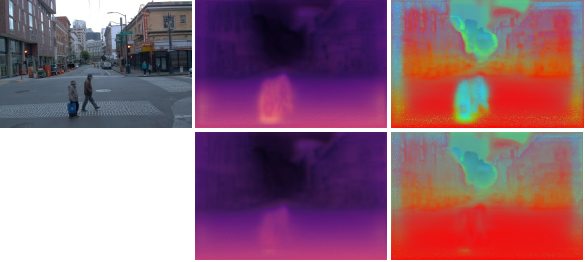}~
    \includegraphics[width=0.49\linewidth]{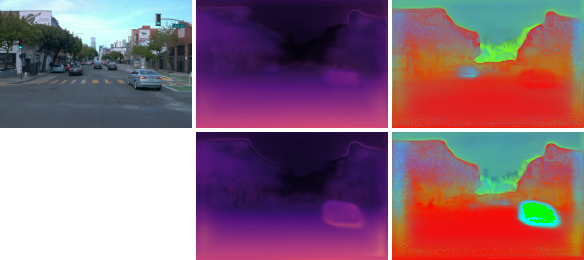}%
    %\vspace{-6pt}
    \caption{
    Depth and motion-segmentation predictions from the first frame of four clips from \waymo.
    For each clip we show the input frame, the 3D structure (as a depth map; brighter is nearer), and the motion segmentation (different colors correspond to different motion components, \ie comoving entities).
    The rows are 2--3 different samples from our model (\ie different draws of $z$); we see the model predicts diverse motion segmentations, where different cars and pedestrians are either static or moving.
    %While the motion segmentation varies from sample to sample, 
    Predicted depths are similar across samples---as expected since depth is much less ambiguous than motion given just one frame.
    }
    \label{fig:waymo-dfs}
\end{figure}

\section{Experiments}
\label{sec:experiments}

\paragraph{Datasets.}
We evaluate our model on two challenging datasets of real-world videos.
Waymo Open Perception~\cite{sun2020cvpr} (\waymo) contains 1150 videos of length 20s at 10FPS, filmed by an autonomous car's sensor suite\footnote{This dataset is licensed for research use and available at \url{https://waymo.com/open/}}.
We use the main front-facing camera only, and the official train/validation/test splits.
As this dataset is relatively small, for final evaluation on the test split, we train on the union of the train and val splits, keeping the hyperparameters fixed.
We downsample the frames to $384 \times 256$ to use as input for our model, and use 6-frame clips.
RealEstate10K~\cite{zhou18siggraph} (\rexk) contains approximately 79000 videos of varying duration, typically a few seconds\footnote{This dataset is available at \url{https://google.github.io/realestate10k/}, with no license specified}.
For testing, we use the first 250 videos from the validation split, as there is no test set released.
The original dataset has a frame-rate of 30FPS; we subsample it to 10FPS, which makes the task more challenging as there is more movement from frame to frame.
We downsample then center-crop the frames to $384 \times 256$, and use 12-frame clips.
For both datasets, we use the provided camera intrinsics (\ie focal length and principal point).
During training we sample clips randomly from within each video; we also apply data augmentation: random horizontal flipping, and random perturbations to contrast and color (see Appendix~\ref{app:augmentation} for details).
During testing we take just one clip per video, starting at the first frame.
We emphasize that the chosen datasets have significant camera motion and perspective effects, in contrast to datasets often used to evaluate video prediction that have a static camera and background~\cite{KTH,ionescu14pami,ebert17corl}.

\newcommand{\stdev}[1]{\footnotesize~$\pm$#1}
\begin{table}
  \caption{Quantitative performance on video prediction for our method (\ours) and SRVP~\cite{franceschi20icml} on two datasets. For descriptions of metrics, please see main text. Note that our method outperforms SRVP according to every metric. Small numbers give the standard deviation over three random seeds.
  %, with Ours\textsubscript{veh} being slightly better than Ours\textsubscript{gm} for \waymo
  }
  \label{tab:results}
  \vspace{-2pt}
  \centering
  \renewcommand{\arraystretch}{1.2}
  \begin{tabular}{@{}llccccc@{}}
    \toprule
    ~ & ~ & FVD $\downarrow$ & KVD $\downarrow$ & PSNR $\uparrow$ & SSIM $\uparrow$ & LPIPS $\downarrow$ \\
%    \multicolumn{2}{c}{Part}                   \\
%    \cmidrule(r){1-2}
    \midrule
    \waymo & \ours\textsubscript{gm} & 1052\stdev{232} & 188.5\stdev{31.6} & 23.16\stdev{0.52} & 0.648\stdev{0.035} & 0.266\stdev{0.041} \\
    ~ & \ours\textsubscript{veh} & 827\stdev{37} & 160.5\stdev{5.8} & 23.60\stdev{0.15} & 0.673\stdev{0.013} & 0.251\stdev{0.009} \\
    ~ & \ours\textsubscript{6-cpt} & 1015\stdev{120} & 189.4\stdev{21.9} & 22.97\stdev{0.09} & 0.653\stdev{0.015} & 0.278\stdev{0.009} \\
    ~ & SRVP & 2087\stdev{142} & 269.4\stdev{7.3} & 20.84\stdev{0.09} & 0.493\stdev{0.002} & 0.574\stdev{0.037} \\
    \midrule
    \rexk & \ours\textsubscript{gm} & 647\stdev{24} & 51.0\stdev{1.1} & 18.72\stdev{0.32} & 0.536\stdev{0.017} & 0.343\stdev{0.016} \\
    ~ & SRVP & 1163\stdev{14} & 79.6\stdev{0.9} & 16.68\stdev{0.08} & 0.414\stdev{0.003} & 0.594\stdev{0.005} \\
    \bottomrule
  \end{tabular}
\end{table}

\begin{figure}[t]
    \centering
    input\hspace{1.3cm}$t=0.1s$\hspace{1.15cm}$t=0.2s$\hspace{1.15cm}$t=0.3s$\hspace{1.15cm}$t=0.4s$\hspace{1.15cm}$t=0.5s$\\
    \href{http://pmh47.net/vipl4s/#waymofig}{
    \includegraphics[width=\linewidth]{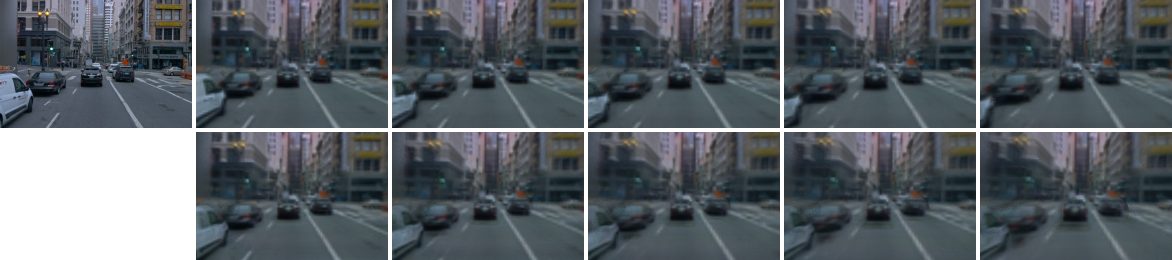}\\
    \includegraphics[width=\linewidth]{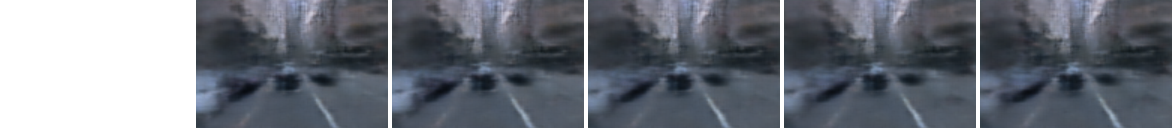}\\[4pt]
    \includegraphics[width=\linewidth]{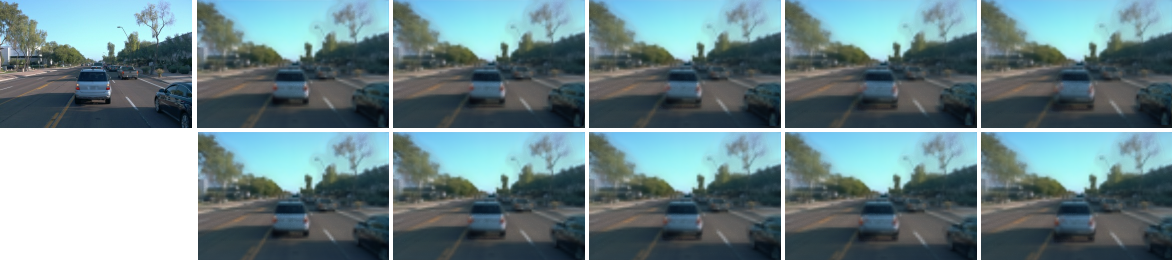}\\
    \includegraphics[width=\linewidth]{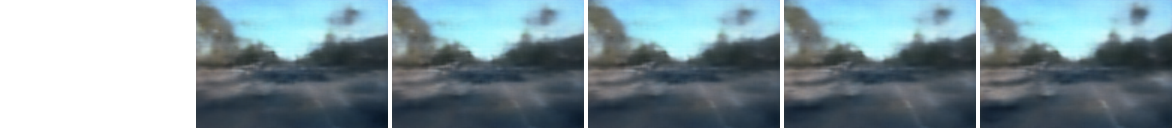}\\[4pt]
    \includegraphics[width=\linewidth]{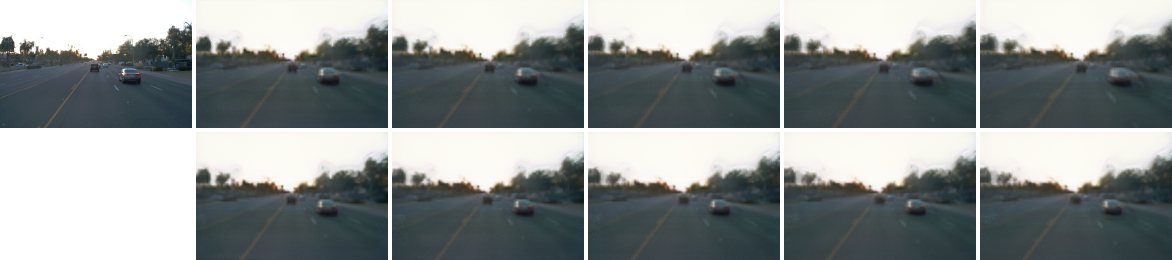}\\
    \includegraphics[width=\linewidth]{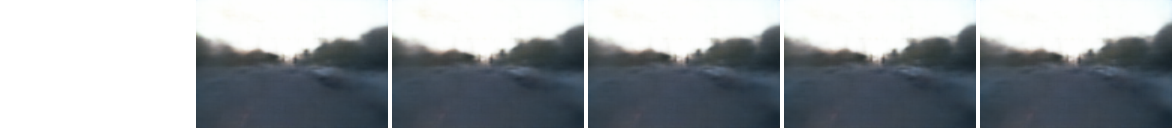}%
    }
    %\vspace{-6pt}
    \caption{Videos predicted by our model and SRVP~\cite{franceschi20icml} given an initial frame from three clips in \waymo. The left column is the input frame; the remaining columns shown different time steps into the future.
    For each input frame, we show two different samples from our model.
    We see that our model (first two rows for each clip) samples diverse yet plausible camera and car velocities---note the changes in relative position among vehicles and background.
    SRVP (third row for each clip) fails to model the motion of the camera through the scene coherently.
    \textit{Best viewed as animations---click the images to link to the project page.}
    }
    \label{fig:waymo-frames}
\end{figure}

\paragraph{Models.}
We name our model \textit{\ours}, short for Video Prediction with Latent 4D Scenes.
On \waymo, we compare three variants of our model.
\ours\textsubscript{gm} uses the general camera motion model (\ref{sec:method}), allowing arbitrary motion.
\ours\textsubscript{veh} uses the more-specialized vehicle motion model, with fewer degrees of freedom in the motion parametrization.
\ours\textsubscript{6-cpt} uses six motion components, instead of four for the previous variants. Thus, the model can separate the scene into more independently-moving regions, which may be beneficial in scenes with a large number of vehicles moving at different velocities.
On \rexk, we only use the general motion parametrization (\ours\textsubscript{gm}), as the videos are shot using handheld cameras and drones.
We also do not experiment with more motion components, as moving objects are exceedingly rare in this dataset.
We implemented our model in TensorFlow~\cite{abadi15tensorflow}; the code will be made public soon.
Hyperparameters were set by incremental searches over groups of related parameters; we did not perform an exhaustive search due to computational constraints.
The values used for our experiments are given in Appendix~\ref{app:hyperparams}.
As a baseline, we use the state-of-the-art stochastic video prediction model \textit{SRVP}~\cite{franceschi20icml}.
We adapted their publicly-available implementation\footnote{The code is Apache licensed and available at \url{https://github.com/edouardelasalles/srvp}} slightly to our setting, increasing the encoder and decoder depth, and re-tuned the hyperparameters on our datasets.
Each model was trained on four Nvidia GTX 1080 Ti GPUs (or two for SRVP) on a local compute cluster, until validation-set convergence or for a maximum of five days (whichever was sooner).

\paragraph{Metrics.}
We evaluate performance using five metrics.
\textit{PSNR}, \textit{SSIM}, and \textit{LPIPS}~\cite{zhang18cvpr} are standard image-similarity metrics; to evaluate stochastic video generation, we follow the common protocol~\cite{denton18icml,lee18arxiv,franceschi20icml} of drawing 100 samples from our model for each input, evaluating the metrics on each sample, and taking the best.
These three metrics therefore measure how similar is the closest sample from a model to the ground-truth clip.
Fr\'echet video distance (\textit{FVD})~\cite{unterthiner19iclrw} and kernel video distance (\textit{KVD})~\cite{binkowski18iclr,unterthiner19iclrw} measure how close a \textit{distribution} of generated videos is to the ground-truth distribution; they both operate by passing sets of ground-truth and generated videos through the action-recognition network of \cite{carreira17cvpr}, then measuring divergences of the resulting distributions of features.
These two metrics therefore measure the realism and diversity of outputs from a model, without considering whether the \textit{true} future frames are predicted for each clip.
For a fair comparison with \cite{franceschi20icml}, we sample images at $128 \times 128$ resolution for the quantitative evaluation---the same as for their model.
For all results, we report the mean and standard deviation over three runs with different random seeds.

\begin{figure}[t]
    \centering
    \newcommand{\overlaypreddepth}[1]{\begin{overpic}[width=\linewidth]{#1}\put(5.0,0.8){\textcolor{white}{\scriptsize predicted depth}}\end{overpic}}
    input\hspace{1.3cm}$t=0.2s$\hspace{1.15cm}$t=0.4s$\hspace{1.15cm}$t=0.6s$\hspace{1.15cm}$t=0.8s$\hspace{1.15cm}$t=1.2s$\\
    \href{http://pmh47.net/vipl4s/#re10kfig}{
    \overlaypreddepth{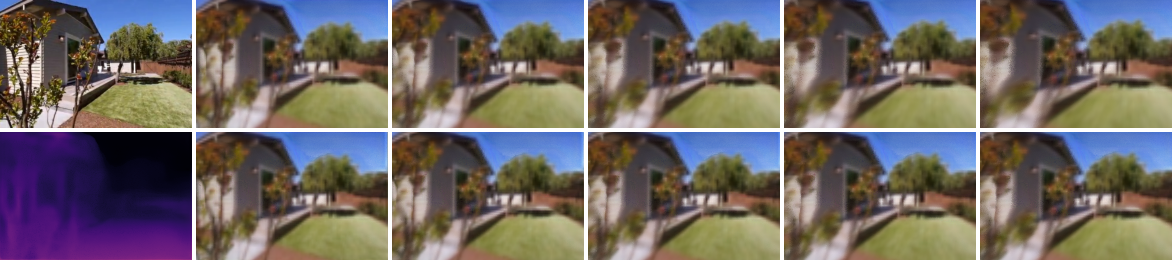}\\
    \includegraphics[width=\linewidth]{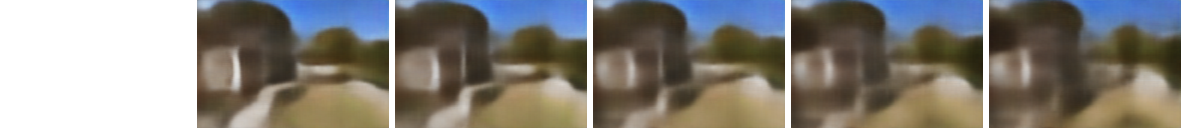}\\[4pt]
    \overlaypreddepth{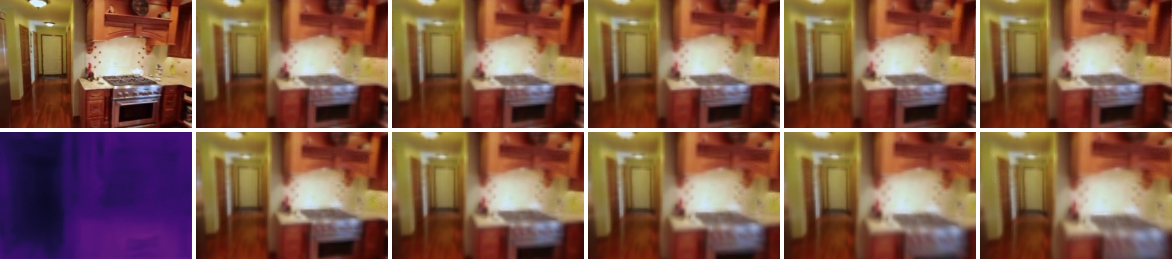}\\
    \includegraphics[width=\linewidth]{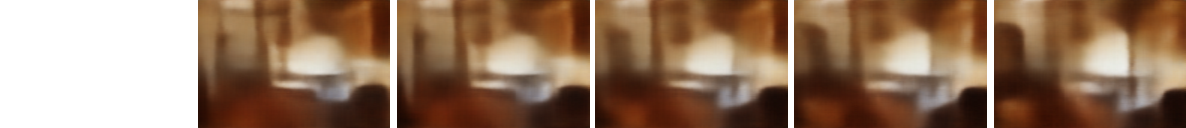}\\[4pt]
    \overlaypreddepth{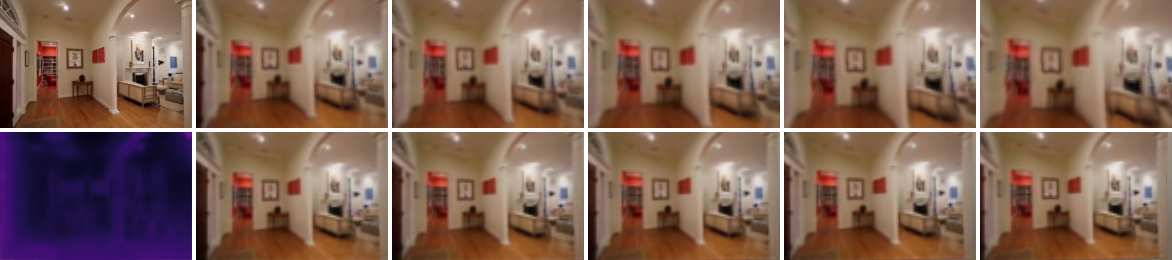}\\
    \includegraphics[width=\linewidth]{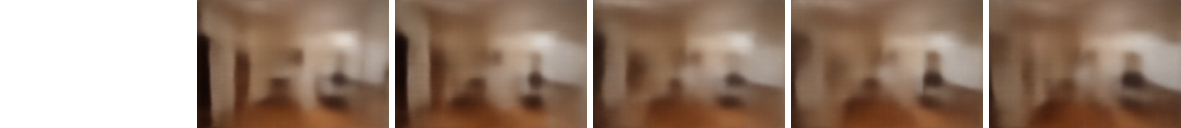}%
    }
    %\vspace{-6pt}
    \caption{Videos predicted by our model and SRVP~\cite{franceschi20icml} given an initial frame from \rexk. 
    We model 12-frame clips, but show only every 2nd frame.
    We also show the depth map predicted by \ours from one sample (brighter is nearer).
    Our model (first two rows for each clip) has learnt a reasonable distribution of camera motion (\ie the camera moves smoothly without passing through objects); the 3D structure represented by the depth-map is less well-defined than for \waymo, particularly in untextured regions.
    SRVP (third row for each clip) fails to capture the textural details of the sequence, nor model the motion coherently.
    \textit{Best viewed as animations---click the images to link to the project page.}
    }
    \label{fig:rexk-frames}
\end{figure}

\paragraph{Results.}
Quantitative results from our method and the baseline are given in Table~\ref{tab:results}.
On both datasets, our general model \ours\textsubscript{gm} outperforms SRVP~\cite{franceschi20icml} across all metrics.
In particular, higher PSNR and SSIM, and lower LPIPS, indicate that the closest samples from our model to the true frames for each clip, are significantly more similar than the closest samples from SRVP.
The difference is significantly smaller for PSNR, likely because this is less sensitive to small-scale and textural details, which we found SRVP predicts poorly.
FVD and KVD are substantially lower for our method, indicating that the distribution of clips generated is also closer to the ground-truth distribution than those generated by SRVP.
Overall, PSNR, SSIM and LPIPS are worse for \rexk than \waymo; this may be due to larger areas of the image being textured than \waymo, which has extensive, smooth sky and road regions (note that FVD and KVD are not directly comparable across datasets, due to the nature of the metrics~\cite{unterthiner19iclrw}).

The two additional variants of our model tested on \waymo, \ours\textsubscript{veh} and \ours\textsubscript{6-cpt} also perform better than SRVP.
However, the 6-component variant is slightly worse than the standard (4-component) variant, despite having strictly greater expressive power.
We hypothesize this is due to optimization difficulties, as we observed for example that the model may reach local optima where several components are used to model different parts of the static background, when one would suffice.
In contrast, the variant with a camera motion parametrization specialized to vehicles performs slightly better than that with a general motion model.
While the differences are within error bars for some metrics, this still provides some support for the hypothesis that incorporating extra prior knowledge helps the generative model learn efficiently.

In Figure~\ref{fig:waymo-dfs}, we show examples of the 3D structure and motion segmentations predicted by our model \ours\textsubscript{veh} for clips from \waymo.
For each input frame, the model samples diverse but plausible motion segmentations.
In particular, it has discovered without supervision that cars are sometimes (but not always) in motion, as are pedestrians.
In Figure~\ref{fig:waymo-frames} we show examples of clips sampled from the same model, and from SRVP~\cite{franceschi20icml}.
Our ability to segment moving cars results in sequences where we can observe relative motion between camera, cars and background, including significant variation between samples for the same input frame.
The corresponding samples from SRVP typically do not exhibit coherent motion, in contrast to those from our model.

Finally, we show sampled clips and 3D structure for \rexk in Figure~\ref{fig:rexk-frames}, for both \ours{} and SRVP.
Here we model longer (12-frame) sequences, but only show every second frame.
The depth-maps are typically accurate around edges, but poor in untextured regions, due to the lack of learning signal here from the reconstruction loss.
Still, the reconstructed clips are plausible; the model has successfully learnt how the camera moves in the training data (\eg it does not pass through walls).

\section{Discussion}
\label{sec:discussion}

\paragraph{Limitations.}
As the first model of its kind to address video prediction via a 4D scene representation and without supervision, our approach currently has several limitations:
\begin{itemize}[itemsep=1pt,leftmargin=16pt,topsep=0pt]
\item It is more computationally-expensive to sample clips at test time than the baseline~\cite{franceschi20icml}, due to the requirement to evaluate the scene function $F$ at many locations along every ray. %Recent work on efficient NeRF evaluation should mitigate this~\cite{?}.
\item We do not incorporate sophisticated inductive biases in the dynamics model (\eg, \cite{minderer19neurips,kossen20iclr,jiang20iclr,lin20icml})---we simply predict the object motion in `one shot' from a densely-connected network.
%
%Indeed, we currently assume the dynamics can be captured by rigid-body transformations; while this is sufficient for many scenarios (\eg the motion components might correspond to rigid parts rather than objects), it cannot model arbitrary deformations without a prohibitively large number of components.
%
\item Ideally, the component assignments $\omega_j$ would be one-hot, but in practice we must relax them to allow gradient-based training. An unfortunate consequence is the introduction of local optima where points are assigned partly to one component and partly to another; this results in `ghosting' artifacts where an object appears to move in two directions at once. %\pmh{do we include any examples to point to?}
\item It can be seen in Figure~\ref{fig:waymo-dfs} that the sky is often assigned to the same motion component as moving foreground objects.
The model does not receive a significant penalty for this, as the sky is both very distant and lacking textural detail, so moving at the speed of a pedestrian or even a car has negligible effect on the resulting frames.
\item We currently assume that the object appearances remain constant over time, whereas in fact the \textit{albedo} should remain so, yet illumination changes. It would therefore be valuable to incorporate recent work on separating these in the NeRF representation~\cite{srinasavan21cvpr}.
\end{itemize}

%Although our method yields state-of-the-art performance on video prediction, its outputs are not yet of such high quality that they can be mistaken for real videos. Hence, for now, we do not consider this work has any negative societal impacts.

\paragraph{Conclusion.}
We have presented a new model \textit{\ours} that can predict video clips from a single frame.
It incorporates an explicit, stable and consistent representation of the 4D scene depicted by the resulting clip.
Moreover, we have seen that on two challenging real-world datasets, it can sample diverse and plausible futures, with differing object and camera motion.
Finally, we showed that its quantitative performance significantly exceeds that of an existing state-of-the-art method that cannot exploit 3D structure~\cite{franceschi20icml}.

\begin{ack}
This research was supported by the Scientific Service Units (SSU) of IST Austria through resources provided by Scientific Computing (SciComp).
\end{ack}

{
\small

\bibliographystyle{plain}
\bibliography{shortstrings,references}
}

\appendix

\section{Network Architectures}
\label{app:architectures}

Here we describe the encoder and decoder network architectures for each component of our model.
Unspecified parameters are assumed to take Keras defaults.

\subsection{Variational encoder $\mathrm{enc}_\phi$}

\begin{tabular}{l}
    \toprule
        Downsampling3D(factor=[1, 2, 2]) \\
        Conv3D(32, kernel size=[1, 7, 7], strides=[1, 2, 2], activation=relu) \\

        GroupNormalization(groups=4) \\
        Conv3D(48, kernel size=[1, 3, 3], activation=relu) \\
        Conv3D(64, kernel size=[3, 1, 1], activation=relu) \\
        MaxPool3D(size=[1, 2, 2]) \\

        GroupNormalization(groups=8) \\
        Conv3D(48, kernel size=[1, 3, 3], activation=relu) \\
        Conv3D(64, kernel size=[3, 1, 1], activation=relu) \\
        MaxPool3D(size=[1, 2, 2]) \\

        GroupNormalization(groups=8) \\
        Conv3D(48, kernel size=[1, 3, 3], activation=relu) \\
        Conv3D(64, kernel size=[2, 1, 1], activation=relu) \\
        MaxPool3D(size=[1, 2, 2]) \\

        Conv3D(128, kernel size=[1, 3, 3], activation=relu) \\
        GroupNormalization(groups=8) \\
        Flatten \\
        Dense(512, activation=relu) \\
        Residual(Dense(activation=relu)) \\
        Dense($128 \times 2$) \\
    \bottomrule
\end{tabular}
\vspace{1em}

\subsection{Scene function $F$}

\begin{tabular}{l}
    \toprule
        Dense(192, activation=relu) \\

        $
        \left.
        \begin{array}{@{}l}
        \text{Dense(192)} \\
        \text{LayerNormalization} \\
        \text{FiLM} \\
        \text{LeakyReLU} \\
        \end{array}
        \right\} \times 3 \text{~with residual connections}
        $ \\

        Dense($1+3+J$) \\
    \bottomrule
\end{tabular}
\vspace{1em}

\subsection{Conditioning Network}

\subsubsection{U-net}

\begin{tabular}{l}
    \toprule

        Downsampling2D(factor=2) \\
        Conv2D(48, kernel size=[3, 3], activation=relu) \\
        GroupNormalization(groups=8) \\
        MaxPool2D(size=[2, 2]) \\

        Conv2D(64, kernel size=[3, 3], activation=relu) \\
        GroupNormalization(groups=8) \\
        MaxPool2D(size=[2, 2]) \\

        Conv2D(96, kernel size=[3, 3], activation=relu) \\
        GroupNormalization(groups=8) \\
        MaxPool2D(size=[2, 2]) \\

        Conv2D(128, kernel size=[3, 3], activation=relu) \\
        GroupNormalization(groups=8) \\
        MaxPool2D(size=[2, 2]) \\

        Conv2D(192, kernel size=[3, 3], activation=relu) \\
        GroupNormalization(groups=8) \\
        MaxPool2D(size=[2, 2]) \\

        Conv2D(192, kernel size=[3, 3], activation=relu) \\
        \textit{concatenate with latent $z$ to give $\xi'$} \\

        Upsampling2D(factor=2), \\
        \textit{concatenate with skip connection} \\
        Conv2D(192, kernel size=[3, 3], activation=relu) \\
        GroupNormalization(groups=8) \\

        Upsampling2D(factor=2), \\
        \textit{concatenate with skip connection} \\
        Conv2D(128, kernel size=[3, 3], activation=relu) \\
        GroupNormalization(groups=8) \\

        Upsampling2D(factor=2), \\
        \textit{concatenate with skip connection} \\
        Conv2D(96, kernel size=[3, 3], activation=relu) \\
        GroupNormalization(groups=8) \\

        Upsampling2D(factor=2), \\
        \textit{concatenate with skip connection} \\
        Conv2D(64, kernel size=[3, 3], activation=relu) \\
        GroupNormalization(groups=8) \\

    \bottomrule
\end{tabular}
\vspace{1em}

\subsubsection{$\xi'$ to $\xi$}

\begin{tabular}{l}
    \toprule
        Conv2D(256, kernel size=[3, 3], activation=relu) \\
        GroupNormalization(groups=16) \\
        Flatten \\
        Dense(1024, activation=relu) \\
        LayerNormalization \\
        Residual(Dense(activation=relu)) \\
    \bottomrule
\end{tabular}
\vspace{1em}

\subsubsection{$\xi$ to FiLM conditioning}

\begin{tabular}{l}
    \toprule
        Residual(Dense(activation=relu)) \\
        LayerNormalization \\
        Residual(Dense(activation=relu)) \\
        LayerNormalization \\
        Residual(Dense(activation=relu)) \\
        LayerNormalization \\
    \bottomrule
\end{tabular}
\vspace{1em}

\subsubsection{$\xi$ to component transformation parameters}

\begin{tabular}{l}
    \toprule
        Dense(512, activation=relu) \\
        Dense($(L-1) \times J \times 2$, kernel initializer=zeros) \\
    \bottomrule
\end{tabular}
\vspace{1em}

\subsubsection{$\xi$ to camera transformation parameters (\ours\textsubscript{gm})}

\begin{tabular}{l}
    \toprule
        Dense(512, activation=relu) \\
        Dense($(L-1) \times 5$, kernel initializer=zeros) \\
    \bottomrule
\end{tabular}
\vspace{1em}

\subsubsection{$\xi$ to camera transformation parameters (\ours\textsubscript{veh})}

\begin{tabular}{l}
    \toprule
        Dense(512, activation=relu) \\
        Dense($(L-1) \times 2$, kernel initializer=zeros) \\
    \bottomrule
\end{tabular}
\vspace{1em}

\section{Camera Transformation Parametrizations}
\label{app:camera-transforms}

As noted in the main text (Section~2), we consider two different parametrizations for the camera transformations $\Lambda_t$.
The first allows arbitrary translation and rotation, while the second is specialized to cameras mounted on vehicles.

\paragraph{General motion.}
We map the embedding $\xi$ to five values per frame, representing $xyz$ translation, yaw (azimuth) and pitch (elevation).
These values are interpreted as absolute values of the parameters (\ie not velocities or changes from the previous frame).
We apply the corresponding transformations to the camera in the order pitch, then yaw, then translation.
We assume camera roll is zero, which is the case for all our datasets, and that the full transformation in the initial frame is the identity.

\paragraph{Vehicle motion.}
For this parametrization, we assume that all camera motion occurs in the $xz$ (ground) plane.
We map the embedding $\xi$ to two values per frame, representing forward speed $s_t$, and azimuthal velocity $\alpha_t$.
The absolute azimuth $a_t$ at frame $t$ is then given by $\sum_{t'=1}^t \alpha_{t'} s_{t'}$; note that $\alpha_t$ is multiplied by the forward speed, so when the vehicle is stationary (zero speed), it is also not allowed to be rotating.
The linear velocity $v_t$ at frame $t$ is given by $(\sin a_t,\, 0,\, -\cos a_t) \cdot s_t$, and the position by $\sum_{t'=1}^t v_{t'}$.

%\section{Object Transformation Parametrization}

\section{Regularizers}
\label{app:regularizers}

As noted in the main text, we use several regularizers to prevent degenerate solutions.
We pretrain the unsupervised single-frame depth-prediction method of \cite{li2020corl} on our data, yielding approximate depth-maps $d_t$ for every input frame.
We also extract keypoint tracks from the input frames using the self-supervised method \cite{detone18cvpr}.
Then when reconstructing frame $x_t$ during training, we render 2D flow and segmentation maps $\mathcal{F}_t$ and $\mathcal{M}_t^j$.
These are calculated similarly to the RGB pixels $x_t$, as in Eq.~(2)--(4) of the main text, but replacing the colors $c(\cdot)$ with the flow (from the initial frame to frame $t$) and motion component indicators respectively.
Given these quantities, we define the following regularizers:

\begin{itemize}
    
    \item L1 regularization on the velocities of motion components, \ie translations between successive component transformations $T_t^j$:
    \begin{equation}
        \frac{1}{J} \sum_j \frac{1}{L} \sum_t \left\lVert T_t^j(\mathbf{0}) - T_{t-1}^j(\mathbf{0}) \right\rVert_1
    \end{equation}
    
    \item Edge-aware TV-L1 regularization~\cite{BoykovICCV01,li2020corl,henderson20neurips} on the component masks $\mathcal{M}_t^j$, penalizing boundaries between motion components for occurring in areas where the ground-truth image $x_t^*$ has small spatial gradients.
    Specifically, let $K_G$ be a $5 \times 5$ Gaussian smoothing kernel, and $D_x , D_y$ be central-difference derivative kernels; for each reconstructed frame $t$ and motion component $j$, we then minimize
      \begingroup\makeatletter\def\f@size{9}\check@mathfonts
      \begin{equation}
        \hspace{-6pt}
        \int_\Omega
        \left\{
          \left| D_x * \mathcal{M}_t^j \right| \exp \hspace{-2pt} \left( -\zeta \left| D_x * K_G * x_t^* \right| \right)
          +
          \left| D_y * \mathcal{M}_t^j \right| \exp \hspace{-2pt} \left( -\zeta \left| D_y * K_G * x_t^* \right| \right)
        \right\}
        \mathrm{d} q
      \end{equation}\endgroup
      where $q \in \Omega$ ranges over pixels in a frame and $\zeta$ is a hyperparameter.

    \item We regularize $\rho$ to be consistent with $d_t$, by first grouping sample locations $r_t^k$ according to whether they lie within, nearer, or further than a slab around $d_t$ of thickness $d_t \times 0.025$.
    For points that are nearer, we regularize them to have near-zero density, by minimizing $R[p-0.01]$
    where $R[\cdot]$ is ReLU, and 
    \begin{equation}
        p=1-\prod_j \left\{ 
            1- \left( 1-\exp\left\{-\rho\big((T_t^j)^{-1}[r^t_k]\big)\right\} \right) \, \omega_j\big((T_t^j)^{-1}[r^t_k]\big)
        \right\}
    \end{equation}
    is the probability of a ray terminating in unit distance near the sample location.
    For points that are within the slab, we regularize them to have large density, by minimizing $R[0.975-p]\times 6.5 / W$, where $W$ is the total number of such points and $6.5$ is an empirically-determined weighting factor. Note that we normalize by $W$ in this case but not for points nearer than the slab, as we only require one point within the slab (\ie near the predicted depth) to have high density.
    
    \item L2 regularization on the difference between the predicted flows $\mathcal{F}_t$ and the displacements along keypoint tracks.
    Specifically, for each frame $t$ and each keypoint that appears in both the initial frame (at location $k_0$) and the $t$\textsuperscript{th} frame (at location $k_t$), we minimize $\lVert (k_t-k_0) - \mathcal{F}_t(k_0) \rVert$ where $\mathcal{F}_t(k_0)$ gives the image-space predicted flow from the initial frame to frame $t$, at location $k_0$ in the initial frame.
    
    \item L2 regularization on the difference between the true keypoint locations and reprojected locations based on $d_t$ and $\Lambda_t$. Specifically, for each frame $t$ and each keypoint that appears in both the initial frame (at location $k_0$) and the $t$\textsuperscript{th} frame (at location $k_t$), we calculate $\tilde{k}_t^j = \pi_t T_t^j \pi_0^{-1}k_0 \,\forall j$, where $\pi_t$ is the projective transformation mapping a 3D point to image-space according to the camera transformation $\Lambda_t$; then we minimize $\min_j \lVert k_t - \tilde{k}_t^j \rVert$. Thus, we require that the motion of the camera and one of the components is consistent with the observed motion of the keypoint.
    
\end{itemize}

\section{Data Augmentation}
\label{app:augmentation}

We apply the following data augmentation during training, uniformly to whole clips:
\begin{itemize}
    \item Random horizontal flipping, with probability $\frac{1}{2}$
    \item Random additive perturbation of HSV color, mapping $(h, s, v)$ to $(h + \delta h,\,s + \delta s,\,v + \delta v)$ with $\delta h \in [-0.05,\,0.05]$, $\delta s \in [-0.2,\,0.1]$, $\delta v \in [-0.1,\,0.1]$, followed by clamping to $[0,\,1]$
    \item Random contrast adjustment by a factor $f \in [0.8,\,1.2]$, mapping component $x$ of each pixel to $(x - \mu) \cdot f + \mu$, where $\mu$ is the mean of that channel across all pixels
\end{itemize}

\section{Hyperparameters}
\label{app:hyperparams}

Here we list all hyperparameters for our model \ours\textsubscript{gm}.
Note that our other model variants \ours\textsubscript{veh} and \ours\textsubscript{6-cpt} use the same hyperparameters with the exceptions described in the main text, \ie \ours\textsubscript{6-cpt} sets $J=6$.

~

\begin{tabular}{lc}
    \toprule
    \textbf{Model} \\
    dimensionality of $z$ & 128 \\
    component count $J$ & 4 (\waymo) / 1 (\rexk) \\
    image size & $384 \times 256$ \\
    \textbf{Loss} \\
    pixel standard deviation & 0.085 \\
    KL weight $\beta$ & 1 \\
    KL annealing steps & 50000 \\
    L1 velocity strength & 0.1 \\
    edge-aware TV-L1 strength & 10 \\
    edge-aware TV-L1 $\zeta$ & 10 \\
    $d_t$ to $\rho$ consistency strength & 100 \\
    keypoint to $\mathcal{F}_t$ consistency strength & 2 \\
    keypoint to $d_t$ consistency strength & 500 \\
    \textbf{Optimization} \\
    batch size & 8 \\
    learning rate & $10^{-4}$ \\
    \bottomrule
\end{tabular}
\vspace{1em}

\noindent The following lists those hyperparameters for the baseline SRVP~\cite{franceschi20icml} which we changed from their default values.

~

\begin{tabular}{lc}
    \toprule
    image size & $128 \times 128$ \\
    batch size & 120 \\
    dimensionality of $y$ & 64 \\
    dimensionality of $z$ & 64 \\
    pixel standard deviation & 0.05 \\
    conditioning frames & 1 \\
    sequence length & 6 (\waymo) / 12 (\rexk) \\
    \bottomrule
\end{tabular}
\vspace{1em}

\end{document}